\RequirePackage{fix-cm}
\documentclass[smallextended]{svjour3}       % onecolumn (second format)
\smartqed  % flush right qed marks, e.g. at end of proof

\usepackage{amsmath,amssymb}
\usepackage{color}
\usepackage{subfig}
\usepackage{epsfig}

\usepackage{multirow}
\begin{document}

\title{Robust Tracking via Weighted Online Extreme Learning Machine%\thanks{Grants or other notes
}


\author{Jing Zhang   \and
        Huibing  Wang\and
        Yonggong Ren*  %etc. \
}


\institute{Jing Zhang\at
              School of computer and information technology, Liaoning Normal University, Dalian, Liaonig, China  116081\\
              \email{zhangjing\_0412@163.com}           %  \\
           \and
           Huibing Wang\at
              Information Science and Technology College, Dalian Maritime University, Dalian, Liaoning, China 116024 \\
              \email{whb08421005@mail.dlut.edu.cn}
           \and
           Yonggong Ren\at
              School of computer and information technology, Liaoning Normal University, Dalian, Liaonig, China  116081\\
              \email{ryg@lnnu.edu.cn}
}

\date{Received: date / Accepted: date}

\maketitle

\begin{abstract}
The tracking method based on the extreme learning machine (ELM) is efficient and effective. ELM randomly generates input weights and biases in the hidden layer, and then calculates and computes the output weights by reducing the iterative solution to the problem of linear equations. Therefore, ELM offers the satisfying classification performance and fast training time than other discriminative models in tracking. However, the original ELM method often suffers from the problem of the imbalanced classification distribution, which is caused by few target objects, leading to under-fitting and more background samples leading to over-fitting. Worse still, it reduces the robustness of tracking under special conditions including occlusion, illumination, etc. To address above problems, in this paper, we present a robust tracking algorithm. First, we introduce the local weight matrix that is the dynamic creation from the data distribution at the current frame in the original ELM so as to balance between the empirical and structure risk, and fully learn the target object to enhance the classification performance. Second, we improve it to the incremental learning method ensuring tracking real-time and efficient. Finally, the forgetting factor is used to strengthen the robustness for changing of the classification distribution with time. Meanwhile, we propose a novel optimized method to obtain the optimal sample as the target object, which avoids tracking drift resulting from noisy samples. Therefore, our tracking method can fully learn both of the target object and background information to enhance the tracking performance, and it is evaluated in 20 challenge image sequences with different attributes including illumination, occlusion, deformation, etc., which achieves better performance than several state-of-the-art methods in terms of effectiveness and robustness.
\keywords{Target tracking \and Extreme learning machine \and  Online sequential extreme learning machine \and Imbalance dataset}
\end{abstract}

\section{Introduction}
\label{intro}
With the rapid development of Internet, multi-media, sensors, more and more image data generate at all time. Target tracking as an important method of image processing has been applied in various fields, including robot navigation, 3D reconstruction, medical diagnosis and so on. However, target tracking suffers from challenging problems due to the video surveillance image sequences usually containing little clarity and fuzzy, which are caused by climate situation, illumination, occlusion, etc. Therefore, how to enhance the tracking performance becomes a hot issue with focusing from domestic and foreign scholars. The target tracking includes usually three steps: target description, target detection, target optimization. First, in order to reduce the influence of redundancy and noisy information, the feature extracting method is used to describe the target object. Second, the discriminant model is exploited to obtain the candidate set from in the stage of target detection. Finally, achieve the optimal target object from the candidate set at the current frame. Target optimization avoids tracking drift that is caused by noisy samples, and target detection is the prerequisite of target optimization. Therefore, target detection and target optimization are important steps in tracking. How to obtain robust and efficient target detection and optimization method become an urgent problem.
\par Recently, feature extracting models are used to enhance the tracking performance including histogram equalization [1], gray level transformation [2], retinex [3] and so on. Of these, histogram equalization expresses color features of images and the gray value distribution effectively. However, the histogram feature of image has less sensibility to rotate, motion, zoom. For the above problem, Bala, etc. proposed structure element method [4] to express texture feature of image achieving better performance. However, information of between extracted interest points and image is mismatching in this method, and structure element has a strong dependence for images. Therefore, the research that exploits structure element to express image in tracking is limited. In order to obtain the global describe of images, Revaud etc. take advantage of geodesic distance of real interpolation to extract shape feature of image [5]. However, the dimensional of the extracted feature is rather high, and products redundant information. For this problem, multi-scale feature extraction based on compressed sensing had been developed [6, 7]. The method that utilizes relatively small randomly generated linear mapping achieves more accurate the original information of images. Zhang, etc. is inspired by compressed sensing, and proposed CT method in 2012 [8]. CT method not only fully consideration tracking drift problem, but also well keep the original structure of image. Therefore, this method obtains better performance in target tracking. Meanwhile, Zhang, etc. improved CT model to present FCT model that employs non-adaptive random projections that preserve the structure of the image feature space of objects and compress sample images of the foreground target and the background using the same sparse measurement matrix. This method reduces the computational complex and obtains the real-time tracking performance [9].
\begin{figure}
\begin{center}
\begin{tabular}{c}
\includegraphics[height=2.2cm]{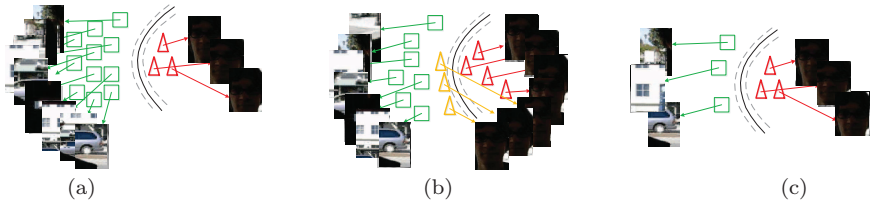}  % fig2 includes two images
\\
(a) \hspace{4.1cm} (b) \hspace{4.1cm} (c)
\end{tabular}
\end{center}
\caption
{ \label{fig:example2}
Example of imbalance dataset classification: (a) the classification boundary closing to the small number of positive samples in training, (b) positive samples are misclassification resulting from under-fitting, (c) removing negative samples to obtain balance dataset leading to lack of environment information.}
\end{figure}
\par Above methods enhance the tracking performance from the view of the target description. However, target tracking are continuous and dynamic, the target object at the current frame generates the influence to follow-up tracking directly. Therefore, achieving the suitable detecting method contributes to avoid tracking drift rather effectively. The detecting methods based on discriminative models define the target tracking as binary-classification problem, and then classify between the target object and background. Meanwhile, obtain the classifier that is used at the next frame. Classification methods are used to build tracking detection such as Naive bayes [8, 9], SVM [10], Adaboost [11, 12], ELM [13], and so on. However, target tracking is different from traditional the classification problem, as the tracking performance is influence by the environment information strongly. In order to solve occlusion, motion blur, background clutter, etc., environment and background of image sequences need to learn adequately. However, in some methods, the number of positive samples is usually less than negative samples, for instance, in OAB algorithm [11] only one positive sample is used in tracking. At this situation, the dataset that is consisted by positive and negative samples is imbalance. Original detecting methods based on discriminative will generate under-fitting to the target object and over-fitting to background samples, thereby reduce the detecting accuracy. As shown in Fig. 1(a), the classification boundary closes to positive samples in training process. However, in Fig. 1(b) (best view in color), yellow samples are failed classified in testing process, which result from the under-fitting to positive samples. Traditional methods consider that remove redundant negative samples reconstruct balance dataset. This method will leads to tracking detection losing efficacy for complex background image sequences. As shown in Fig. 1(c), we remove many negative samples, and then less environment information leads to ineffective classifiers. For above problem, we propose a robust tracking method. First, we achieve a classification model based on the dynamic weight matrix for imbalance samples. Second, improve it to the incremental model, and it is applied in tracking detection at each frame. Finally, we propose an optimization method to obtain the optimal sample as the target object from the candidate set. The model framework is displayed in Fig. 2.
\par The major contributions of the proposed approach are as follows:
\begin{enumerate}
  \item We present a robust tracking model based on ELM. This model achieves the satisfying performance in target tracking under special conditions including deformation, occlusion, etc. Meanwhile, it is an incremental tracking approach, so it obtains the real-time tracking.
  \item The proposed model fully considers the attitude of the imbalance dataset in target detecting, and exploits the local weight matrix to avoid both problems that are under-fitting to target objects and over-fitting to background samples. Meanwhile, introduces the forgetting factor to improve the robustness for changing of the classification distribution with time.
  \item In order to enhance the robustness and avoid tracking drift, we propose the novel optimization method to obtain the optimal target object from the candidate set at each current frame.
\end{enumerate}
\par This paper is organized as follows. We firstly review the most relevant work on tracking in Section 2. The preliminaries of ELM and OSELM are introduced in Section 3. The proposed model and discussion are detailed in Section 4, and the discussion in Section 5. The experimental results are presented in Section 6 with comparisons to state-of-the-art methods on challenging sequences. The conclusion and our future work in Section 7.
\section{Related work}
\label{sec:1}
Tracking detection based on building approaches of models are divided to generative tracking and discriminative tracking. Generative tracking models exploit the minimum error of reconstruction to describe the target object. Meanwhile, search the most-similar region from image sequences to obtain the target object, which classical methods including mean-shift [14] and sub-space [15]. Mean-shift utilizes a similar function to metric the distance between different images. Sub-space models regard image as the high-dimension matrix that is composed of pixels, and video is composed of the continuous image sequence that is correlation between images. In [16], Chrysos, etc. proposed the offline sub-space model to enhance the tracking performance. Ross, etc. proposed the incremental sub-space method to learn the original model in real-time, and it overcomes the deformation that is caused by rotating and illuminating [17]. However, it is difficult to achieve the robust performance when the above method under the complex background. With the study on sparse representation deeply, some scholars utilize sparse representation to build tracking models. Mci etc. presented the l1-tracker model that regards the target object as sparse representation of the target object template and trivial template to deal with the occlusion problem [18]. However, the computational complexity of the l1-tracker is rather high, thereby Li, etc. further improved this method by using the orthogonal matching pursuit for enhancing the efficiently [19]. Meanwhile, in order to improve the efficiently, Bao, etc. utilized the accelerated proximal gradient method to solve the optimization problem [20]. To address the optimization of L1, Liu and Sun, etc. presented the tracking model based on reverse sparse representation that represents a static template of the target object by using particle and occlusion template to construct dictionary [21]. To address the similar object problem in tracking, Jia, etc. presented the ASLA model based on a novel alignment-pooling [22]. Above tracking methods based on generative models only describe the appearance of models, and ignore the information of environment and background. The accuracy and stability of tracking performance depends on a great extent on divisibility of target objects and background samples. Therefore, it is difficult of generative tracking to achieve the satisfying performance when the target in complex situations.
\par Discriminative tracking methods regard the detecting problem as the binary classification problem that builds the discriminative model utilizing between of the target object and background samples. Avidan etc. exploited SVM joining on the optical flow method to in tracking [23]. Compared to the SVM classification model, Boosting is a simple and fast method to construct the classifiers relatively, and it is more appropriate to solve image process problems. Parag, etc. established the tracking model based on Boosting framework [24]. Zhang, etc. regarded Naive Bayes as the classifier in tracking detection [8]. Moreover, Zhang, etc. proposed STC model that formulates the spatio-temporal relationships between the interested object and its locally dense contexts in the Bayesian framework to solve target ambiguity effectively [25]. However, the separability will changes as the appearance of the target object and background information make changing in tracking. For above problems, classifiers based on discriminative models are enhanced to incremental learning in tracking. Colline, etc. maximized the variance ration of between the target object and the background to obtain features with the high identifiability in tracking [26]. Grabner, etc. proposed the online-updating Adaboost algorithm that is used in tracking [27]. And then, Babenko, etc. presented the incremental learning model based on multi-sample [28].Unsupervised classification methods are widely applied in the image processing field due to without the help of any the label information. Wang, etc. proposed representative studies including [29-43]. Kalal, etc. introduced the semi-supervised Adaboost model in the tracking framework [44], which improves the tracking performance dramatically. Wang, etc. proposed super-pixel method to solve problems of the occlusion and the cover [45]. Above the tracking detection methods based on discriminative models enhance the tracking performance by optimizing the classification accuracy. Recently, the deep learning model have been proposed and applied in face recognition, segment and so on. This model processes input data directly to obtain features of images, and establishes the discriminative model. This model is quite superior in terms of its parallel computing, nonlinear mapping, self-learning and so on. Li, etc. took advantage of the CNN model to solve target tracking [46]. Zhou, etc. exploited deep network composing of online boosting to enhance the tracking performance [47]. These method based on deep learning exploit only one image at the firstly frame to construct training dataset would lead to the size of training dataset that is small and appear the problem of under-fitting. To address above problems, Hunag, etc. proposed ELM model that is single hidden level neural network [48-50]. This model convert solve-iterate into linear equation solving by random setting hidden parameters. Comparing to deep learning model has many advantages including simpler network structure, without a large number of the training dataset, therefore, enhance the speed of solving and avoid trapping into local optimal solution. In order to generalize effectiveness of ELM, many scholar present improved algorithms [51-54]. Meanwhile, ELM is applied in various fields and obtains the satisfying performance. Zhang, etc. utilized the incremental learning model of ELM to enhance the tracking robustness and effectiveness [55].
\section{Preliminaries}
The proposed novel method is based on ELM. In order to facilitate the understanding of our method, this section briefly reviews the related concepts and theories of ELM and developed OSELM [52].

\subsection{ELM}
\label{sect:title}
\par Extreme learning machine is improved by single hidden layer neural network (SLFNs): assume given $N$ samples $(X,T)$, where $X=[x_{1},x_{2},...,x_{N}]^{T} \in\mathbb{R}^{d\times N}$, $T=[t_{1},t_{2},...,t_{N}]^{T} \in\mathbb{R}^{\tilde{N}\times N}$, and $t_{i}=[t_{i1},t_{i2},...,t_{im}]^{T} \in\mathbb{R}^{m}$. The method is used to solve multi-classification problems, and thereby the number of network output nodes is $m(m\geq 2)$. There are $\widetilde{N}$ hidden layer nodes in networks, and activation function $h(\cdot)$ can be Sigmoid or RBF: $\sum_{i=1}^{\widetilde{N}} \beta_{i}h(a_{i}x_{j}+b_{j})=o_{j}$
where $j=1,\cdots,\tilde{N}$, $a_{j}=[a_{j1},a_{j2},\cdots,a_{jd}]^{T}$ is the input weight vector, and $\beta_{j}=[\beta_{j1},\beta_{j2},...,\beta_{jm}]^{T}$ is the output weight vector. Moreover, $a_{j}$, $b_{j}$ can be generated randomly, which is known by [47, 48]. Written in matrix form: $H\beta =T$, where $H_{i} = [h_{1}(a_{1}x_{1}+b_{1}),\cdots, h_{N}(a_{\widetilde{N}}x_{1}+b_{\widetilde{N}})]$. Moreover, the solution form of $H\beta=T$ can be written as: $\hat{\beta}=H^{\dag}T$, where $H^{\dag}$ is the generalized inverse matrix of $H$. ELM minimize both the training errors and the output weights. The expression can be formulated based on optimization of ELM:
\begin{equation}
\begin{array}{c}
  \rm Minimize: \it L_{ELM}=\frac{1}{2}\|\beta\|^{2}_{2}+C\frac{1}{2}\sum_{i=1}^{N} \|\xi_{i}\|^{2}_{2}\\
  \\
  \rm Subject \; to: \it t_{i} \beta\cdot h(x_{i})\geq 1-\xi_{i},i=1,...,N\\
  \\
  \xi_{i}\geq 0,i=1,...,N
\end{array}
\end{equation}
where $\xi_{i}=\left(
                 \begin{array}{ccc}
                   \xi_{i,1} & \cdots & \xi_{i,m} \\
                 \end{array}
               \right)
$ is the vector of the training errors. We can solve the above equation based on KKT theory by Lagrange multiplier, and can obtain the analytical expression of the output weight: $\hat{\beta} = H^{T}(\frac{I}{C}+HH^{T})^{-1}T$. The output function of ELM is: $f(x)=h(x)\hat{\beta}=h(x)H^{T}(\frac{I}{C}+HH^{T})^{-1}T$.
\par We can know that ELM expect to minimize the structural risk, equivalent to maximize the distance from samples to the classification boundary. However, in order to minimize the structural risk, the classification boundary is close enough to small number of samples when the classification distributing of samples is imbalance, liking Fig.1. Therefore, product the under-fitting problem to small number samples, and incline generalization of the model. Aiming to above problem, in (1) utilize adding weight matrix to adjust the structural risk. According to equation (1), and we are inspired by [51] induce the weight matrix $W$, and obtain the improved optimization equation, where each line of $W$ correspond a training sample.
\begin{equation}
\begin{array}{c}
  \rm Minimize: \it L_{ELM}=\frac{1}{2}\|\beta\|^{2}_{2}+CW\frac{1}{2}\sum_{i=1}^{N} \|\xi_{i}\|^{2}_{2}\\
  \\
  \rm Subject \; to: \it t_{i} \beta\cdot h(x_{i})\geq 1-\xi_{i},i=1,...,N\\
  \\
  \xi_{i}\geq 0,i=1,...,N
\end{array}
\end{equation}

\par According to KKT condition, we can obtain the output weight expression as follow:
\begin{equation}
\hat{\beta}=(\frac{I}{C}+H^{T}WH)^{-1}H^{T}WT
\end{equation}
\subsection{OSELM}
The above model is used to solve classification problem for static batch data. Aiming to this problem Rong et al. proposed an increment classification model OSELM [51]. It is an online solving algorithm based on ELM. The model trains the $\Delta N(\Delta N\geq 1)$ chunk of new samples to obtain new model, then uses matrix calculation with the original model. Through the above calculation, the new output weight matrix
$\hat{\beta} _{N + \Delta N}$ is obtained. When the new $\Delta N$ chunk arrives, the hidden output weight matrix is updated. The expression is listed as follows:
\[{H_{N + \Delta N}} = \left[ {\begin{array}{*{20}{c}}
{h({x_1};{a_1},{b_1})}& \cdots &{h({x_1};{a_{\tilde N}},{b_{\tilde N}})}\\
 \vdots &{}& \vdots \\
{h({x_N};{a_1},{b_1})}& \cdots &{h({x_N};{a_{\tilde N}},{b_{\tilde N}})}\\
{h({x_{N + 1}};{a_1},{b_1})}& \cdots &{h({x_{N + 1}};{a_{\tilde N}},{b_{\tilde N}})}\\
 \vdots &{}& \vdots \\
{h({x_{N + \Delta N}};{a_1},{b_1})}& \cdots &{h({x_{N + \Delta N}};{a_{\tilde N}},{b_{\tilde N}})}
\end{array}} \right] = \left[ {\begin{array}{*{20}{c}}
{{h_1}}\\
 \vdots \\
{{h_N}}\\
{{h_{N + 1}}}\\
 \vdots \\
{{h_{N + \Delta N}}}
\end{array}} \right] = \left[ {\begin{array}{*{20}{c}}
{{H_N}}\\
{{H_{\Delta N}}}
\end{array}} \right]\]
where
${h_{N + k}} = {\left[ {\begin{array}{*{20}{c}}
{h({x_{N + k}};{a_1},{b_1})}& \cdots &{h({x_{N + k}};{a_{\tilde N}},{b_{\tilde N}})}
\end{array}} \right]^T}(k = 1, \ldots ,\Delta N)$ is the $k$th new sample corresponding the vector. Therefore, the output vector is
${T_{\Delta N}} = {\left[ {\begin{array}{*{20}{c}}
{t_{N + 1}^{}}& \cdots &{t_{N + \Delta N}^{}}
\end{array}} \right]^T}$.
Therefore, the incremental expression of the output weight is obtained:
\begin{equation}
{\hat{\beta} _{N + \Delta N}} = {(H_N^T{H_N} + H_{\Delta N}^T{H_{\Delta N}})^{ - 1}}(H_N^T{T_N} + H_{\Delta N}^T{T_{\Delta N}})
\end{equation}
Let ${G_0} = {(H_N^T{H_N})^{ - 1}}$, and the incremental expression $G_1$ can be written as:
\begin{equation}
{G_1}^{ - 1} = {G_0}^{ - 1} + H_{\Delta N}^T{H_{\Delta N}}
\end{equation}
\par According to the equation (4) and (5), the new output weight matrix $\hat{\beta} _{N + \Delta N}$ becomes:
\begin{equation}
{\hat{\beta} _{N + \Delta N}} = {G_1}({H_N}{T_N} + H_{\Delta N}^T{T_{\Delta N}}) = {\beta _N} + {G_1}H_{\Delta N}^T({T_{\Delta N}} - {H_{\Delta N}}{\beta _N})
\end{equation}
where ${G_1} = {(G_0^{ - 1} + H_{\Delta N}^T{H_{\Delta N}})^{ - 1}}$.
\par According to the above equation, we formulate the further expression:
\begin{equation}
{G_1} = {G_0} - {G_0}H_{\Delta N}^T{({I_{\Delta N}} + {H_{\Delta N}}{G_0}H_{\Delta N}^T)^{ - 1}}{H_{\Delta N}}{G_0}
\end{equation}
From the above learning process, OSELM trains new model by adjusting the original model according to the equation (7) when the new dynamic samples are arriving.

\section{Proposed algorithm}
In this paper, the proposed tracking model mainly contains three steps: sampling and sample representation, tracking detection, optimization of the candidate set.

\subsection{Sampling and samples representation}
In this paper, we assume that the tracking window had been labeled by handle in the first frame. Meanwhile, we obtain positive and negative samples by sampling in fixed range. In order to ensure positive samples are similar with the target object, the Euclidean distance is used to filter sampled samples. Meanwhile, the sampling method needs to meet two conditions: 1) Positive samples closer to the target object at the current frame. 2) Negative samples faster to the target object at the current frame, and negative samples contain more information of environment and background. The specific process of sampling is as follows:
\begin{enumerate}
  \item  The target object at the current frame is taken as the center of sampling range to collect positive and negative samples. We record the top-left corner coordinate $(xP, yP)$ (See Figure 3(a),(b)), the width and height of the sliding window. The range of the positive area is: $[x{P_{\min }},y{P_{\min }}],[x{P_{\max }},y{P_{\max }}]$, and the negative area is: $[x{N_{\min }},y{N_{\min }}], [x{N_{\max }},y{N_{\max }}]$.
  \item The sliding window slithers in a fixed area with the step $wD$ to obtain positive and negative samples.
  \item Euclidean distance $sD$ is used to filer further all samples, and $sD$ (See Figure 3(c)) is calculated from the center of the target object to all of samples. The filter conditions is $[r_{1},r_{2}],[r_{3},r_{4}]$.
  \item According to the distance $sD$, determines the classification of each collected sample. If $sD$ satisfies $r_1^2 < sD < r_2^2$, the corresponded sample belong with the positive classification. Otherwise, if $sD$ satisfies $r_3^2 < sD < r_4^2$, the corresponding samples belong with negative classification.
\end{enumerate}

\begin{figure}
\begin{center}
\begin{tabular}{c}
\includegraphics[height=3cm]{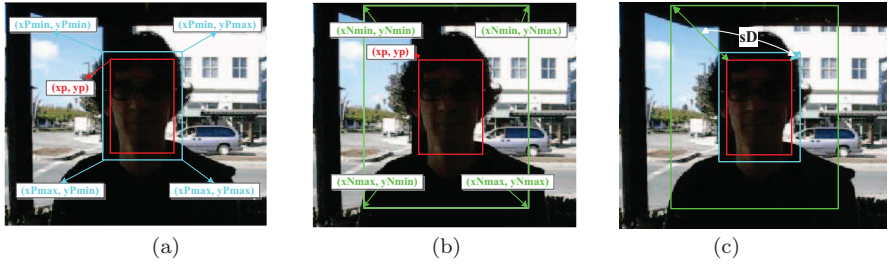}  % fig2 includes two images
\\
(a) \hspace{3.1cm} (b) \hspace{3.1cm} (c)
\end{tabular}
\end{center}
\caption
{ \label{fig:example2}
 The area of samples and the Euclidean distance: (a) the area of positive samples, (b) the area of negative samples, (c) the Euclidean distance.}
\end{figure}

This sampling process is shown in Figure 2 and Figure 3. The sample set is established and provides the prerequisite of training and tracking for follow-up.
\par According to above steps, we obtain positive and negative samples that are used in tracking detection. Meanwhile, we can know that the sampled sample set is imbalance dataset from the sampling process. Therefore, if take advantage of traditional classification method will produce under-fitting to the small number samples and over-fitting to the large number samples.
\par In order to enhance the expressiveness of features, and reduce the time-consume in tracking. The dimensionality reduction method is used to extract feature of samples. In this section, we utilize the random projection theory to extract features of samples from [14]. The random matrix $P$ is used to map $x\in\mathbb{R}^d $ from high dimensional space to low dimensional space $q\in\mathbb{R}^v$:
\begin{equation}
q = Px
\end{equation}
where $v\ll d$. By the Johnson-Lindenstrauss lemma theory, high probability the distances between
the points in a vector space are preserved, if they are projected onto a randomly selected subspace with suitably high dimensions. Meanwhile, Baraniuk et al. [31] proved this theory, and applied it in the compressed sensing. Therefore, if the projection matrix $P$ satisfies the Johnson-Lindenstrauss lemma condition, the low dimensional representation of $x$ can be obtained, where $P \in\mathbb{R}^{N \times d},{p_{i,j}} \sim N(0,1)$. $P$ is a dense matrix, thus it has high computation complexity. We will transform it into a sparse random metric matrix, and expression is listed as follows:
\begin{equation}
{p_{i,j}} = \sqrt \tau \times \left\{ {\begin{array}{*{20}{c}}
{\begin{array}{*{20}{c}}
1&{with}&{probability}&{1/2\tau}
\end{array}}\\
{\begin{array}{*{20}{c}}
0&{with}&{probability}&{1/\tau}
\end{array}}\\
{\begin{array}{*{20}{c}}
{ - 1}&{with}&{probability}&{1/2\tau}
\end{array}}
\end{array}} \right.
\end{equation}
Zhang et al [14] cite and analyze sparsity and computational complexity of $P$ when $\tau=2,3$. And select $\tau=m/4$ to generate a more sparse random matrix. At this point, in each row of $P$ only about $c,c\leq 4$, thus the computational complexity is O(CN) that is very low. Meanwhile, only the non-zero items of the matrix are considered in this algorithm, which makes the algorithm has less space complexity. In order to maintain the scale invariance of the image features, the Haar transform is performed on each image sample. For each sample $x$ has a convolution operation with the filter set, and the multi-scale rectangular filter set is $\{ {l_{1,1}}, \ldots ,{l_{w,h}}\}$, where ${l_{i,j}}$ is defined as:
\begin{equation}
{l_{i,j}}(x,y) = \left\{ {\begin{array}{*{20}{c}}
{1,1 \le x \le i,1 \le y \le j}\\
{0,otherwise}
\end{array}} \right.
\end{equation}
where $i,j$ is the width and height of the rectangular filter, respectively. The filter is generated only at the first frame, as shown in Figure 2. After filtering, features of the samples are extracted using equation (9) and (10).
\subsection{Target detection via weight online extreme learning machine}
Most target detecting methods based on discriminant, these methods learn a binary classification model and then incremental train to update model at the next frame. According to the Eq. (6), OSELM updates the existent model using the current samples to achieve rapid incremental classification. However, OSELM is difficult to solve the incremental classification problem for imbalance dataset. Aiming to this problem we propose WOSELM algorithm in tracking detection.
\par In order to maintain the decomposability of output weight $\hat{\beta}$ when incremental learning, we combine the Eq. (3) to split the matrix $W$. $W$ is a diagonal matrix, thus we can obtain $W=\sqrt{W}\cdot\sqrt{W}$. The output weight change as following form:
\begin{equation}
\hat{\beta}=(\frac{1}{2}+H^{T}WH)^{-1}H^{T}WT=[\frac{I}{C}+(\sqrt{W}H)^{T}(\sqrt{W}H)]^{-1}(\sqrt{W}H)^{T}(\sqrt{W}H)
\end{equation}
Recently, we define $A=\sqrt{W}\cdot H, B=\sqrt{W}\cdot T$, and then the equation can change into $\hat{\beta}=(\frac{I}{C}+A^{T}A)^{-1}A^{T}B$.  According to the incremental learning process of OSLM, we can obtain:
\begin{equation}
\hat{\beta}_{N+\Delta N}=(\frac{I}{C}+A_{N+\Delta N}^{T}A_{N+\Delta N})^{-1}A_{N+\Delta N}^{T}B_{N+\Delta N}
\end{equation}
, where $A_{N+\Delta N}^{T}A_{N+\Delta N}=A_{N}^{T}A_{N}+A_{\Delta N}^{T}A_{\Delta N}$. We define $K_{0}=\frac{I}{2C}+A_{N}^{T}A_{N}$, therefore further obtain:
\begin{equation}
K_{1}=\frac{I}{2C}+K_{0}+A_{\Delta N}^{T}A_{\Delta N}
\end{equation}
According to above expression (13), and $A_{N+\Delta N}^{T}B_{N+\Delta N}=A_{N}^{T}B_{N}+A_{\Delta N}^{T}B_{\Delta N}$. We can get as the following expression:
\begin{equation}
A_{N+\Delta N}^{T}B_{N+\Delta N}=K_{0}K_{0}^{-1}A_{0}^{T}B_{0}+A_{1}^{T}B_{1}=K_{0}\beta_{0}+A_{1}^{T}B_{1}
\end{equation}
And $K_{0}=-\frac{I}{2C}+K_{1}-A_{\Delta N}^{T}A_{\Delta N}$, we further obtain:
\begin{equation}
A_{N+\Delta N}^{T}B_{N+\Delta N}=K_{1}\beta_{N}-\frac{I}{2C}\beta_{N}-A_{\Delta N}^{T}A_{\Delta N}\beta_{0}+A_{\Delta N}^{T}B_{\Delta N}
\end{equation}
Because $\beta_{\Delta N}=K_{1}^{-1}[A_{N}\quad A_{\Delta N}][B_{N}\quad B_{\Delta N}]^{T}$, combine to above equations (12) and (15), we can get the final the incremental expressions of the output weight:
\begin{equation}
\begin{aligned}
\hat{\beta}_{N+\Delta N}&=K_{1}^{-1}[K_{1}\beta _{N}-\frac{I}{2C}\beta_{N}-A_{\Delta N}^{T}A_{\Delta N}\beta_{N}+A_{\Delta N}^{T}B_{\Delta N}]\\ &=\beta_{N}-K_{1}^{-1}[K_{1}\beta _{N}-\frac{I}{2C}\beta_{N}-A_{\Delta N}^{T}A_{\Delta N}\beta_{N}+A_{\Delta N}^{T}B_{\Delta N}]
\end{aligned}
\end{equation}
According to the Eq. (16), we can note that not only achieve incremental learning model, but also enhance the classification performance for imbalance dataset. Therefore, how to build the local weight matrix $W$ become an important problem. In this paper, we are inspired by [55] using C-S algorithm that is proposed by Elkan to dynamic generate the local weight matrix $W$. Shown in Fig. 4, we know that $P$ and $N$ are true class, $T$ and $F$ are predicted class. $C_{00},C_{11}$ are cost of corrected classification. $C_{10},C_{01}$ are cost of misclassification. Meanwhile, it is that the small number of samples are misclassified hard accepted by ours. Therefore, $C_{10}>C_{01}$ and $C_{00}=C_{11}$, we can obtain the following expression:
\begin{equation}
p(j=0|x)C_{10}+p(j=1|x)C_{11}\geq p(j=0|x)C_{00}+p(j=1|x)C_{01}
\end{equation}
, where $x$ is sample, $j=0, 1$ is classification label. C-S theory consider that a better classifier making values in left and right side of the Eq. (17) are equal, obtaining as following expression:
\begin{equation}
p(j=0|x)C_{10}+p(j=1|x)C_{11}=p(j=0|x)C_{00}+p(j=1|x)C_{01}
\end{equation}
The proposed method to solve the binary classification problem, thus $p(j=0|x)=1-p(j=1|x)$, and combine the Eq. (18). We can get the following expression:
\begin{equation}
p(j=0|x)=\frac{C_{01}}{C_{01}+C_{10}}
\end{equation}
, where the meaning of $C_{10}$ is cost of false negatives. When all negatives are miss classified, the value of $C_{10}$ equals the number of negatives $P$, similarly, the value of $C_{01}$ equals the number of positives $N$. We obtain the probability of letting $x$ as $P$. Therefore, the weight of $x$ is as following:
\begin{equation}
W_{10}=\frac{N}{P+N}
\end{equation}
Meanwhile, the corresponding $W_{01}=1-W_{10}$ , we can obtain the local weight matrix $W$. The proposed algorithm that is applied in imbalance dataset not only enhances the performance, but also realizes the incremental learning. However, in dynamic imbalance dataset, the solved matrix $W$ that is local weight matrix is difficult to adapt changing of the distribution with time. Show in Fig. 5(a) is the distribution of offline samples. In Fig. 5(b) red samples is new samples, if adjust the classification boundary according to new samples, will produce deviation of the boundary position. In Fig. 5(c), the classification boundary is correct position.
\begin{figure}
\begin{center}
\begin{tabular}{c}
\includegraphics[height=2.2cm]{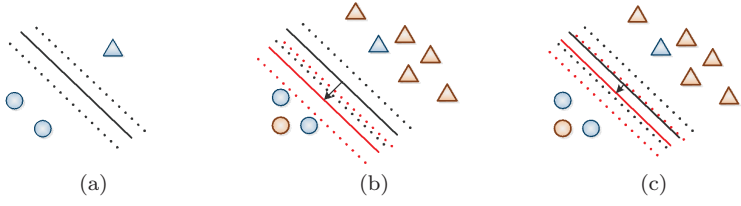}  % fig2 includes two images
\\
(a) \hspace{3.1cm} (b) \hspace{3.1cm} (c)
\end{tabular}
\end{center}
\caption
{ \label{fig:example2}
 Distribution change of imbalance dataset with time: (a) the distribution at previous frame, (b) black line is the faulty classification boundary when new samples arrived and red line is perfect classification boundary, (c) adjusting the classification boundary to adapt new samples.}
\end{figure}
Aiming to this problem, we induce the forgetting factor $\rho$ to weak the existed classifier. We can obtain the classifier at current frame, multiplies it with the forgetting factor, thereby update to the existed classifier. We utilize the elicitation method to obtain the forgetting factor, and define $\rho = |\frac{S_{P}}{S_{N}}-\frac{P}{N}|$, where $S_{P}, S_{N}$, respectively, is the total number of positives and negatives. The expression of output weight $\hat{\beta}$ isㄩ
\begin{equation}
\hat{\beta}_{N+\Delta N}=K_{1}^{-1}[K_{1}\beta _{N}-\frac{I}{2C}(\rho\cdot\beta_{N})-A_{\Delta N}^{T}A_{\Delta N}(\rho\cdot\beta_{N})+A_{\Delta N}^{T}B_{\Delta N}]
\end{equation}
\subsection{Optimizing the candidate set}
The target detecting problem is different from the traditional classification problem that pays more attention to classification performance rather than the special sample. Therefore, obtaining the most optimal target object from the classified results is an important problem in target tracking. If the noisy is regard as the target object, will appear tracking drift. Therefore, the best sample will produce significant impact on follow-up tracking. Most of testing samples are correctly classified. In this figure, green points belong to positive classification, and two positive samples are more close to the classification boundary, thus their features are similar to negative samples. If above situation is taken as a target detecting problem, we want to select the positive sample that is far from the classification boundary. As the right part of Figure 6, the sample labeled \emph{Better Target} should be used as the target object, in the ideal case. In this paper, we define samples that are far from the classification boundary as \emph{Better Target}, and samples that are close to the classification boundary as \emph{Worse Target}. In this paper, positive samples are even correctly classified. They are close to the classification boundary, and therefore they are still \emph{Worse Target}. As shown in Fig. 6, \emph{Worse Target} can not be the target object due to the feature of \emph{Worse Target} approximates with negative samples. Therefore, not only \emph{Worse Target} gets inaccurate detection result, but induces also a huge impact for the samples collection and classification model training at the next frame.

\par We propose an efficient and quick method to address the above-mentioned problem. We utilize the equation (21) to calculate the hidden-layer output weights matrix $\hat \beta$. $H$ was obtained by filtering. According to $H{\hat \beta}={T}(H\in\mathbb{R}^{N\times d}, {\hat \beta}\in\mathbb{R}^{\tilde{N}\times m}, T\in\mathbb{R}^{N\times m})$, the dimension of $T$ is the number of categories. The position of the maximum value of each row of $T$ is the label of the testing sample. According to above analysis, the position relationship is an important factor that affects the tracking performance. Therefore, the matrix $T$ is utilized to judge the position relationships between testing samples and the classification boundary. Specific steps are summarized as follows:
\begin{enumerate}
  \item Calculate the maximum value of each row of $T$ by operation $tmaxV_{i} = max(t_{i})$, and the maximum vector $maxV$ from all of testing samples can be obtained, where $maxV = [tmaxV_{1}, tmaxV_{2},\ldots, tmaxV_{N} ]^{T}\in\mathbb{R}^{N}$.
  \item Sort $maxV$ by using the function $smaxV = sort(maxV)$ from the least to the greatest.
  \item Obtain the sample sequence $xS$ from testing samples in accordance with elements order of $smaxV$.
\end{enumerate}
We define  \emph{Former-samples}. There are ranked ahead in the $xS$ sequence, and are close to the classification boundary more. Meanwhile, defined  \emph{Latter-samples} are ranked behind in the $xS$ sequence, and are far from the classification boundary more. Therefore, we should choose \emph{Latter-samples} as candidates that are \emph{Better Target} in target detecting.
\par However, according to the sampling process in section 4.1, the obtained samples in the current frame has a directly relation with the training samples at the next frame. In other word, if the target sample is the noisy sample at the current frame, it would generates a huge impact on sampling and training at next frame, and event would products visual drift. As shown in Fig. 7, the red circle sample is classified as positive classification, but it is the negative sample and is a noisy sample. According to the above-mentioned target detecting method, the sample that is the farthest distance from the classification boundary would be detected as the target object, but the samples is noisy (\emph{Worse Target}). In this paper, we utilize a novel method that is similar to clustering to solve the problem. The specific describes as follow:
\begin{enumerate}
  \item Get the vector $cN$ from \emph{Latter-samples} of $smaxV$.
  \item Vote for each value $cn_{i}, cn_{i}\in cN$ according to the count of $cn_{i}$.
  \item Obtain the target object that is the largest count of votes in all $cn_{i}$.
\end{enumerate}
Therefore, the sample of the least votes is noisy. In this method, we cluster all \emph{Latter-samples} according to votes. Because the feature of \emph{Worse Target} are different from other features of \emph{Better Target}. Therefore, in order to eliminate the difference of the specific value of the vector $smaxV_{i}$, we only reserved four decimal points. Meanwhile, aiming to reduce the consumed time, the hash structure is used to store the voting process.

\section{Discussion}
According to the description in the above method, we note that this method using the local weight matrix enhances the tracking performance, and then we improve it to incremental learning method to ensure the tracking in real-time. Moreover, we propose the optimizing candidate set method that avoids the tracking drift and enhances the robustness.
\begin{description}
  \item[(1)] Difference with related work. Our method based on discriminating model is a robust tracking method. However, comparing to other tracking methods based on discriminative models, such as OAB, SemiB, CT, FCT, SCT, Struct, our method sampling more background image as negative samples, which enhances the adaptation and robustness of model to adapt changing of background. Meanwhile, unnecessarily consider the adverse effect from imbalance dataset. CT, FCT sampling around the target object, thus only obtain a little background information. It is difficult of these methods to maintain the sensibility to background changing, thereby reduces the tracking performance.
      \par The proposed method obtains the better performance than OAB and SemiB. These reasons can be conclusion as following factors. First, OAB and SemiB utilize Adaboost as classifier, which the iterative computation increases consuming of time. Our method comparing to OAB, SemiB ensures the tracking in real-time. And then, OAB exploits only one sample as the positive sample, so it is difficult to sufficient training of discriminating. Moreover, if the target object is the noisy sample, will results from important tracking drift. The proposed method obtains several positive samples at each frame to construct the candidate set, and introduces the local weight matrix $W$ to adjust the under-fitting and over-fitting that are aroused by imbalance dataset. Therefore, our method achieves more robust and efficient tracking performance.
  \item[(2)] Robustness to ambiguity in detection. Because tracking is different from traditional binary classification problem, it results from the ambiguity problem easily in the multi-positive samples tracking methods. How to seek the best target object from the candidate set becomes an important problem that influences follow-up tracking. To address the above problem and enhancing the robustness of the model, we proposed an efficient method that uses the exponential expression of 2-norm of $maxV$ to obtain the optimal sample as the target object. As shown in Fig. 6 and Fig. 7, we seek the sample that is far away the classification boundary as the target object, and 2-norm of $maxV$ correlated with the distance from samples to the boundary positively. We introduce the exponential expression of $maxV$ 2-norm to enhance the discrimination of between samples. The proposed method exploits the simple operation to optimize the target object, therefore in ensuring real-time tracking with the premise of the robustness.
  \item[(3)] Robustness to occlusion. The proposed method introduces the local weight matrix to solve the imbalance problem, and let the discriminative model can sufficient learning to difference classification samples. Therefore, our method has more attention to the information of background, and strengthens the classifier to recognize the target object. However, traditional tracking detection methods due to lack training to background information, so these difficult to recognize the target object when it is occluded by background image.
  \item[(4)] Real-time tracking. The proposed method through updates the original model at current frame to incremental learning, thereby enhances the calculation speed. The calculation of the local weight matrix is only related to the number of samples, which the solving process is fast process. Meanwhile, in the process of choosing the best target object, we utilize HASH-MAP to realize clustering, which is to further enhance the efficiency of the proposed tracking.
\end{description}

\section{Experimental Results}
The proposed algorithm is run on an i7 Quad-Core machine with 3.4GHz CPU, 16GB RAM and a MATLAB implementation. The average efficient of our method is 30 frame per second(FPS), so this method can achieve the tracking effect in real-time. In order to analyze and compare the strength and weakness of the proposed method comprehensively, we exploit 20 image sequences with different attributes to verify the proposed method. Table 1 summarizes all sequences in detail, where attributes record challenges in tracking including low resolution (LR), in-plane rotation (IPR), out-ofplane rotation (OPR), scale variation (SV), occlusion (OCC), deformation (DEF), background clutters (BC), illumination variation (IV), motion blur (MB), fast motion (FM), and outof-view (OV). In addition, we compare the performance with popular tracking methods including OAB, SemiB based on Adaboost classification model, CT, FCT, STC based on Navie Bayes classification model, Struck based on SVM classification model. Comparisons are summarized in Tab. 2, and Fig. 8-12.
\begin{table}
\caption{Description of image sequence in experiments. }
\label{tab:1}       % Give a unique label
\begin{tabular}{lllll}
\hline\noalign{\smallskip}
Sequence&Target frame & Color & Feature number & Attributes\\
\noalign{\smallskip}\hline\noalign{\smallskip}
Boy & 602 & Yes & 480*640 & IPR, OPR, SV, MB, FM\\
Cardark & 393 & Yes & 240*320 & BC, IV\\
Coke & 291 & Yes & 480*640 & IPR, OPR, OCC, IV, FM \\
Couple & 140 & Yes & 240*320 & OPR, SV, DEF, BC, FM \\
David3 & 252 & Yes & 480*640  & OPR, OCC, DEF, BC \\
Dog & 1350 & No & 240*320 & IPR, OPR, SV \\
Doll & 3872 & Yes & 300*400 & IPR, OPR, SV, OCC, IV \\
Dudek & 1145 & No & 480*720 & IPR, OPR, SV, OCC, IV\\
Faceoccu & 812 & No & 240*320 & IPR, OPR, OCC, IV \\
Fish & 476 & No & 240*320 & IV \\
Girl & 500 & Yes & 96*128 & IPR, OPR, SV, OCC \\
Junmming & 75 & Yes & 226*400 & MB, FM \\
Lemming & 1336 & Yes & 480*640 & OPR, SV, OCC, IV, FM, OV \\
Mhyang & 1490 & No & 240*320 & OPR, DEF, BC, IV \\
Mountbike & 228 & Yes & 360*640  & IPR, OPR, BC \\
Shaking & 365 & Yes & 352*624 & IPR, OPR, BC, SV, IV \\
Singer & 351 & Yes & 352*624 & OPR, SV, OCC, IV \\
SUV & 945 & No & 240*320 & IPR, OCC, IV \\
Trellis & 569 & Yes & 240*320 & IPR, OPR, BC, SV, IV \\
Ironman & 166 & Yes & 304*372 & OPR, SV, BC, IV, MB \\
\noalign{\smallskip}\hline
\end{tabular}
\end{table}

\subsection{Experimental Setup and Evaluation Metrics}
According to the description of the sampling process in Sect. 4.1, the range of the positive classification $[xP_(max), xP_(min)], [yP_(max), yP_(min)]$ is $[-10, 10], [-10, 10]$, and the negative classification $[xN_(max), xN_(min)], [yN_(max), yN_(min)]$ is $[-70, 70], [-70, 70]$. The distance of the sliding window $wD$ is 1. The range of Euclidean Distance is $r1 = 0, r2 = 2, r3 = 8, r4 = 30$. Dimensionality reduction based on CT is used to express the features of image sequences. In this paper, the dimension of each sample is reduce to 50. In the detecting process, the number of the hidden-layer node is 300 in our model.
\par We utilize two kinds of measures to evaluate performances between our method and  state-of-the-art tracking methods. One way for evaluating is the success plot and the precision plot from [57]. We calculate the intersecting area between the target area and the predict area to achieve success plots. And then, get the precision plot in image sequence according to each success plot. The computing formula of the success plot is as follow: $S = Area(B_T \cap B_G)/Area(B_T\cup B_G)$, where $B_T$ is the tracked bounding box and denotes $B_G$ is the truth ground. The success plot shows the percentage of frames with $S>t_0$ throughout all threshold $t_0\in [0, 1]$.
\subsection{Tracking results}
Table 2 summarizes the overall performance of challenging sequences in terms of precision plots, where red fonts indicate the first-best tracking performance in each row of table 2, and green fonts indicate the second-best tracking performance. Meanwhile, OAB, SemiB, STC, CT, FCT, Struck are provided by the author. From results in table 2, we note that our method achieves the first-best performance in most sequences causing by sufficiently considering the influence of environment on tracking. Moreover, the local weight matrix $W$ to adjust the empirical risk avoids the problem that the small number of target objects leads to imbalance. Therefore, our model have satisfying tracking performance, in particular, it obtains the robustness on Suv, Coke, Trellis that are occluded by background. However, FCT attain the first-best tracking performance on Dog, Mountbike, Singer sequence, and CT attain the first-best tracking performance on Duke sequence. OAB and SemiB exploit the Adaboost classification model in detecting process. However, SemiB has the tracking effect is not ideal due to use a lot of unlabel simples to build the discrimination model. STC, CT and FCT utilize the Navie Bayes classification model in tracking detection process. CT and FCT build the low dimension expression and retain important features from original samples, therefore, achieve the second-best performance in challenging sequences. The real-time is an important term of tracking. Our method is an incremental tracking model that only learns the current image and need not to learn the cumulate data in bulk. CT and FCT get the low dimension expression by random matrix, thereby enhance the computation speed. Therefore, FCT is the most effective tracking model among all algorithms. However, in SemiB, the tracking performance is not ideal due to a lot of unlabel samples participating in building the discrimination model. Struck achieves the solution by iterating in the learning process, thus impacts the effectiveness.
\begin{table}
\caption{Success rate(\%)red fonts indicate the first-best tracking performance, and green fonts indicate the second-best tracking performance. }
\label{tab:1}       % Give a unique label
\begin{tabular}{llllllll}
\hline\noalign{\smallskip}
       &Ours & OAB & SemiB & STC & CT  & FCT & Struck \\
\noalign{\smallskip}\hline\noalign{\smallskip}
Boy & {\color{red}89} & {\color{green}73} & 21 & 45 & 48  & 72 & 51 \\
Cardark & {\color{red}80} & 35 & 26 & 18 & 63  & 66 & {\color{green}76} \\
Coke & {\color{red}36} & {\color{green}19} & 0 & 1 & 2  & 1 & {\color{green}19} \\
Couple & {\color{red}65} & 28 & 14 & 0 & 40  & {\color{green}58} & 12 \\
David3 & {\color{red}58} & 42 & {\color{green}43} & 1 & 32  & 10 & 21 \\
Dog & {\color{green}89} & 79 & 49 & 43 & {\color{red}92}  & {\color{green}89} & 66 \\
Doll & {\color{red}91} & 48 & 40 & 10 & 67  & 60 & {\color{green}87} \\
Dudek & {\color{red}76} & 64 & 35 & 51 & {\color{red}76}  & 65 & {\color{green}68} \\
Faceoccu & {\color{red}54} & 38 & {\color{green}43} & 1 & 20  & 42 & 33 \\
Fish & {\color{red}93} & 54 & 1 & 67 & 50  & {\color{green}90} & 72 \\
Girl & {\color{red}59} & 1 & 29 & 1 & 2  & 27 & {\color{green}44} \\
Junmming & {\color{red}60} & 32 & 17 & 29 & 40  & 37 & {\color{green}47} \\
Lemming & {\color{red}68} & 50 & 21 & 18 & 20  & {\color{green}60} & 22 \\
Mhyang & {\color{red}88} & 37 & 33 & 81 & 45  & {\color{green}85} & 63 \\
Mountbike & {\color{green}91} & 63 & 19 & 59 & 18  & {\color{red}95} & 33 \\
Shaking & {\color{red}83} & 1 & 0 & 0 & {\color{green}1}  & {\color{green}1} & {\color{green}1} \\
Singer & {\color{red}75} & 18 & 15 & 22 & 22  & {\color{red}75} & {\color{green}63} \\
SUV & {\color{red}60} & {\color{green}53} & {\color{green}53} & 1 & 24  & 12 & 12 \\
Trellis & {\color{red}28} & {\color{green}1} & 0 & 0 & 0  & 0 & {\color{green}1} \\
Ironman & {\color{red}11} & 4 & {\color{green}6} & 1 & 5  & {\color{green}6} & {\color{red}11} \\
\noalign{\smallskip}\hline
\end{tabular}
\end{table}

\section{Conclusion}
In this paper, we proposed an efficient and robust tracking method based on ELM. The proposed method exploits the local weight matrix to avoid the under-fitting problem for the target object, and we improve it to the incremental learning method to ensure real-time in tracking. Meanwhile, we proposed the optimization method to obtain the optimized target object from the candidate set, thereby avoids the tracking drift problem that is caused by noisy samples. We utilize 20 image sequences to evaluate the proposed method, which achieves more performance than several state-of-the-art methods in effective and robustness. In addition, we will explore more efficient tracking model using the multi-feature image sequences.
\subsubsection*{Compliance with Ethical Standards:}
Funding: This study was funded by the Doctoral Scientific Research Foundation of Liaoning Province (20170520207).

Author Jing Zhang declares that she has no conflict of interest. Author Huibing Wang declares that he has no conflict of interest. Author Yonggong Ren declares that he has no conflict of interest.

Ethical approval: This article does not contain any studies with human participants or animals performed by any of the authors.

Informed consent:  Informed consent was obtained from all individual participants included in the study.

%\begin{acknowledgements}
%\end{acknowledgements}



\end{document}